\documentclass[letterpaper, 10 pt, conference]{ieeeconf}  

\IEEEoverridecommandlockouts                              

\usepackage{cite}
\usepackage{algorithmic}
\usepackage{graphicx}
\usepackage{textcomp}
\usepackage{xcolor}
\usepackage{comment}
\usepackage{amssymb}

\usepackage{amsmath, amsthm}

\usepackage{siunitx}
\usepackage{soul}
\usepackage{multirow}
\usepackage{subcaption}
\usepackage{siunitx}
\usepackage[font=footnotesize]{caption}
\usepackage{hyperref}
\usepackage{mathtools}
\usepackage[ruled,vlined,linesnumbered]{algorithm2e}
\usepackage{bm}
\usepackage[normalem]{ulem}

\usepackage{nomencl}
\usepackage{etoolbox}

\newtheorem{thm}{Theorem}[section]

\newtheorem{assum}[thm]{Assumption}

\newtheorem{rem}{Remark}

\title{\LARGE \bf
Socially Acceptable Bipedal Navigation: A Signal-Temporal-Logic-
Driven Approach for Safe Locomotion
}

\author{Abdulaziz~Shamsah and~Ye~Zhao
\thanks{This work was partially funded by the NSF grants \# IIS-1924978, \# CMMI-2144309, ONR grant \#  N00014-23-1-2223, and Georgia Tech Institute for Robotics and Intelligent Machines (IRIM) Seed Grant.}
\thanks{The authors are with the Laboratory for Intelligent Decision and Autonomous Robots, Woodruff School of Mechanical Engineering, Georgia Institute of Technology, Atlanta, GA 30313, USA
        {\tt\small \{ashamsah3, yezhao\}@gatech.edu}}%
\thanks{Abdulaziz Shamsah is also affiliated with the Mechanical Engineering Department, College of Engineering and Petroleum, Kuwait University, PO Box 5969, Safat, 13060, Kuwait}}

\begin{document}

\maketitle
\thispagestyle{empty}
\pagestyle{empty}

\begin{abstract}

Social navigation for bipedal robots remains relatively unexplored due to the highly complex, nonlinear dynamics of bipedal locomotion. This study presents a preliminary exploration of social navigation for bipedal robots in a human crowded environment. We propose a social path planner that ensures the locomotion safety of the bipedal robot while navigating under a social norm. The proposed planner leverages a conditional variational autoencoder architecture and learns from human crowd datasets to produce a socially acceptable path plan. Robot-specific locomotion safety is formally enforced by incorporating signal temporal logic specifications during the learning process. We demonstrate the integration of the social path planner with a model predictive controller and a low-level passivity controller to enable comprehensive full-body joint control of Digit in a dynamic simulation.

\end{abstract}

\section{Introduction}
\begin{figure}[t]
\centerline{\includegraphics[width=.8\columnwidth]{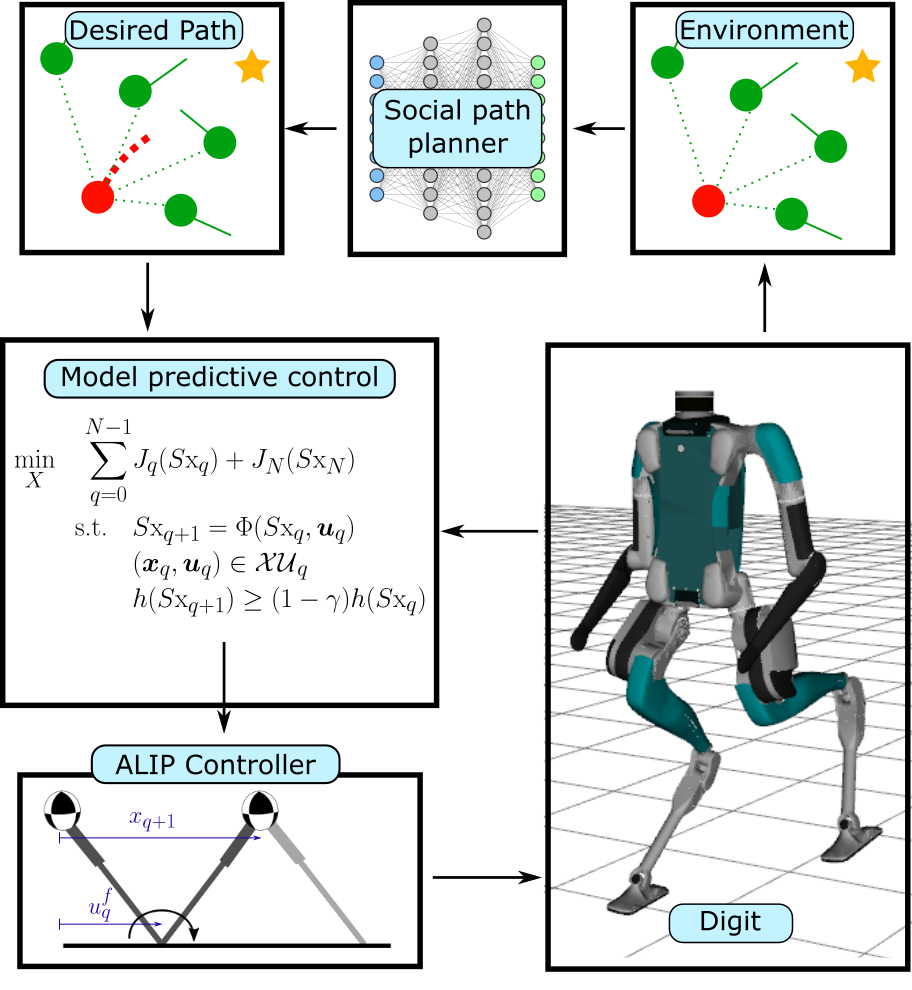}}
\caption{Block diagram of the proposed framework for socially acceptable navigation for the bipedal robot Digit. The social path planner takes the surrounding pedestrians' positions and goal location as input from the environment and then generates a path that gets sent to the Model Predictive Controller (MPC). The MPC solves for desired velocity and heading change commands to the Angular-momentum-based Linear Inverted Pendulum (ALIP) controller to generate foot placements for Digit. }
\label{fig:high-level}
\end{figure}

Bipedal navigation in complex environments has garnered substantial attention within the robotics community~\cite{li2023autonomous, huang2023efficient, narkhede2022sequential, zhao2022reactive, Kulgod2020LTL, warnke2020towards, zhao2017robust}. Very recently, there has been an increasing focus on social navigation for mobile robots in human-life environments~\cite{che2020efficient, bera2019emotionally, paez2022unfreezing, nishimura2020risk, schaefer2021leveraging, majd2021safe}. Nonetheless, the exploration of social navigation in the context of bipedal robots remains largely underexplored. This can be significantly attributed to the nonlinear dynamics challenge associated with bipedal locomotion. In this preliminary study, our objective is to develop a social planner capable of generating robot paths that ensure the safety of the bipedal system Digit~\cite{agility} by integrating locomotion safety into the social planner.

To construct a social path planner, we seek inspiration from the human trajectory prediction community~\cite{mangalam2021goals, mangalam2020not, salzmann2020trajectron++, gupta2018social, sadeghian2019sophie}. The work in \cite{hong2023obstacle} proposes an obstacle avoidance learning method that uses Conditional Variational
Autoencoder (CVAE) framework to learn a temporary target distribution to avoid pedestrians actively. However, during the learning phase, the temporary targets are selected heuristically. In contrast, we aim to learn such temporary waypoints from human crowd datasets. In \cite{mangalam2020not}, the authors introduce both VAE and CVAE architectures to produce accurate pedestrian trajectory prediction, where their learned model is not only conditioned on the past trajectory of the pedestrian but also the intermediate endpoint of the trajectory, representing the pedestrians' intent. 

In contrast to existing pedestrian trajectory prediction models \cite{mangalam2020not, mangalam2021goals, salzmann2020trajectron++} that focus on predicting trajectories of human crowds, our objective is to develop a high-level social path planner specifically designed for bipedal robots. Our learning algorithm is centered around leveraging the robot's sensory data points and locomotion capabilities to enhance its path-planning capabilities. Drawing inspiration from PECNet~\cite{mangalam2020not}, our devised framework adopts a similar CVAE structure but is distinct by taking into account robot-specific locomotion safety and incorporating the ego-agent final destination as the input. As a result, our planner will generate path plans not only imitating how a human would react in a social setting but also providing safety guarantees designed for robotic systems. 

The main contributions of this study are as follows:
\begin{itemize}

\item  High-level social navigation planner for socially acceptable path design utilizing a CVAE architecture.

\item Formally encoding locomotion safety into the high-level learning module. 

\item Showcasing the viability of hierarchically integrating the social path planner with a discrete Model Predictive Controller (MPC) and a low-level passivity controller for full-body joint control of Digit.

\end{itemize}






\section{High-level social path planner}
\label{sec:social_path_planner}


\subsection{Preliminaries}

In this work, we hypothesize that in a social setting, the information accessible by the ego-agent that forms its own future path $(x^{\rm ego}_{[t, t_f]}, y^{\rm ego}_{[t, t_f]}) = \mathcal{T}^{\rm ego}_{[t, t_f]}$ are three folds, first its final destination $(x^{\rm dest}, y^{\rm dest}) =\mathcal{G}$ (intent), the surrounding pedestrians' past trajectory $(x^{p_k}_{[t_p, t]}, y^{p_k}_{[t_p, t]}) = \mathcal{T}^{p_k}_{[t_p, t]}$ for the $k^{\rm th}$ pedestrian, and the ego-agent's social experience, i.e., its own assumptions on how to navigate the environment in a socially-acceptable manner. We treat the social experience as latent information that is not readily available in human crowd datasets. 
Therefore we make the following assumption.


\begin{assum}
Learning the future trajectory of an ego-agent $\mathcal{T}^{\rm ego}_{[t, t_f]}$ based on its final goal $\mathcal{G}$ and surrounding pedestrians' past trajectories $\mathcal{T}^{p_k}_{[t_p, t]}$, will learn the ego-agent's social experience that encompasses implicitly the ego-agent's own prediction of other pedestrians' future trajectories.
\end{assum}

With the intention of implementing the social path planner on our bipedal walking robot Digit~\cite{agility},  $\mathcal{T}^{\rm ego}_{[t, t_f]}$, $\mathcal{T}^{p_k}_{[t_p, t]}$ and $\mathcal{G}$ are all expressed relative to Digit's current position $\mathcal{T}^{\rm ego}_{t}$. Given Digit's limited sensing capability of the surrounding environment, we only consider the pedestrians that are within a radius $r$ at a specific time $t$ and assume that their past trajectories were observable over a specified time interval from $t_p$ to $t$.

\subsection{Learning Architecture}

\begin{figure}[t]
\centerline{\includegraphics[width=.9\columnwidth]{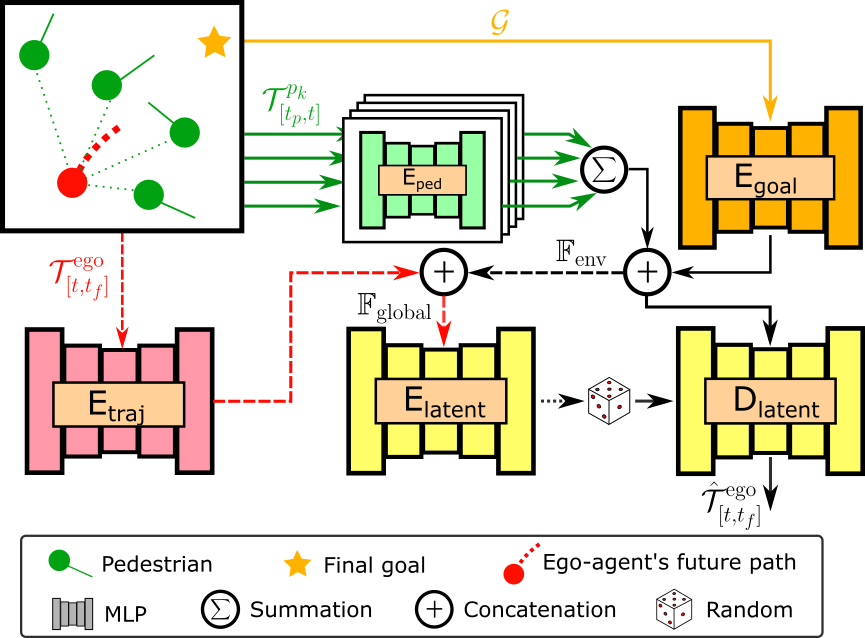}}
\caption{Social path planning module architecture for learning ego agent's trajectory generation conditioned on the relative position of neighboring pedestrians and goal position. Dashed connections are utilized only during training.}
\label{fig:arch}
\end{figure}

Similar to PECNet~\cite{mangalam2020not}, we set up a CVAE architecture to learn the ego-agent's future trajectory conditioned on the final destination goal\footnote{unlike PECNet, we are not seeking to learn the endpoint, as the final destination goal for the robot is known in the navigation task.} and surrounding pedestrians. The proposed architecture incorporates Multi-Layer Perceptrons (MLP) with ReLU non-linearity for all the sub-networks.

The learning phase pipeline is shown Fig.~\ref{fig:arch} and elaborated as follows. The surrounding pedestrians' past trajectories $\mathcal{T}^{p_k}_{[t_p, t]}$ are encoded in $E_{\rm ped}$ individually and the latent features $E_{\rm ped}(\mathcal{T}^{p_k}_{[t_p, t]})$ are then aggregated through summation to take into account the collective effect of the surrounding pedestrian while keeping a fixed architecture\footnote{Other human trajectory learning modules include a social module to take into account the surrounding pedestrians effect such as social non-local pooling mask~\cite{mangalam2020not}, max-pooling~\cite{gupta2018social}, and sorting~\cite{sadeghian2019sophie}.} \cite{salzmann2020trajectron++, ivanovic2018generative, jain2016structural} as seen by the green arrows in Fig.~\ref{fig:arch}.  The goal location for the ego-agent is encoded in $E_{\rm goal}$ as seen by the orange arrow in Fig.~\ref{fig:arch}. The resultant latent features $\sum_{k=1}^{n} E_{\rm ped}(\mathcal{T}^{p_k}_{[t_p, t]})$ and $E_{\rm goal}(\mathcal{G})$ are then concatenated as environment features $\mathbb{F}_{\rm env}$. The ground truth of the ego-agent future trajectory $\mathcal{T}^{\rm ego}_{[t, t_f]}$ is encoded in $E_{\rm traj}$ as shown by the red arrows in Fig.~\ref{fig:arch}. The resultant latent features $E_{\rm traj}(\mathcal{T}^{\rm ego}_{[t, t_f]})$ are then concatenated with $\mathbb{F}_{\rm env}$ as global features $\mathbb{F}_{\rm global}$ and encoded in the latent encoder $E_{\rm latent}$.

We sample potential future trajectories from the latent distribution $\mathcal{N} (\mu,\sigma)$ generated by the $E_{\rm latent}$ module, and concatenate them with $\mathbb{F}_{\rm env}$. This concatenated information is then passed into the latent decoder $D_{\rm latent}$, resulting in our prediction of the ego-agent's future trajectory $\hat{\mathcal{T}}^{\rm ego}_{[t, t_f]}$

\subsection{Formally Incorporating Robot Safety Specifications}
\label{subsec:stl_loss}
\begin{figure}[t]
\centerline{\includegraphics[width=.95\columnwidth]{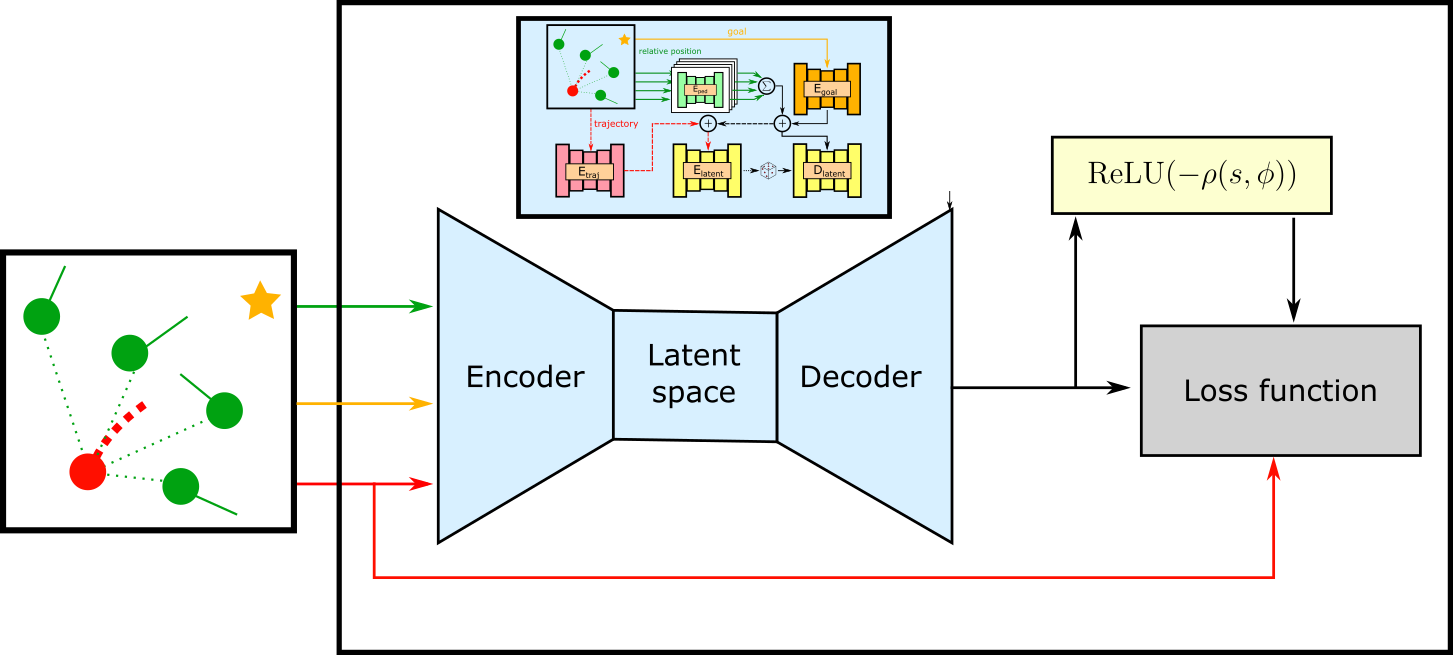}}
\caption{Loss function setup for the neural network with added STL loss}
\label{fig:Stl}
\end{figure}

To deploy the learning-based social path planner on the Digit robot, it is essential to integrate certain desired behaviors. These behaviors are crucial for ensuring robot locomotion safety, but they are not explicitly present in human crowd datasets.

Signal Temporal Logic (STL) is a well-established temporal logic language to formally encode natural language into mathematical representation for control synthesis \cite{maler2004monitoring}. More importantly, the quantitative semantics of STL offer a measure of the robustness of an STL specification $\rho (s_t,\phi)$, i.e., quantify the satisfaction or violation of the specification $\phi$ given a specific signal $s_t$. Positive robustness values indicate satisfaction, while negative robustness values indicate a violation. The authors in \cite{leung2021back} present STLCG, a tool that transforms STL formulas into computational graphs to be used in gradient-based problems such as neural network learning. To this end, we utilize STLCG to formally incorporate desired safety behaviors into our learning framework by encoding STL specifications as additional loss functions that penalize STL formula violation \cite{leung2021back, li2021vehicle}.





The locomotion safety specifications are derived based on our previously introduced Reduced-Order Model (ROM) safety theorems \cite{shamsah2023integrated} and our empirical knowledge about the locomotion safety of Digit \cite{agility} during our experiments. In these specifications, we regulate $\hat{\mathcal{T}}^{\rm ego}_{[t, t_f]}$ to limit the sagittal and lateral Center of mass (CoM) velocities as well as the heading change between consecutive walking steps. 
\subsubsection{Locomotion velocity specification $\phi_{\rm vel}$}
Let $s^{v_{\rm sag}}_{[t+1, t_f]}$ and $s^{v_{\rm lat}}_{[t+1, t_f]}$ be a signal equal to the velocity of $\hat{\mathcal{T}}^{\rm ego}_{[t, t+f]}$ in the sagittal and lateral directions, respectively. The locomotion velocity specification has:
\begin{gather*}
\phi_{\rm sag} = \square_{[t+1, t_f]}(s^{v_{\rm sag}}_{[t+1, t_f]} \leq v_{\rm max} \wedge s^{v_{\rm sag}}_{[t+1, t_f]} \geq v_{\rm min} ) \\
\phi_{\rm lat} = \square_{[t+1, t_f]}(s^{v_{\rm lat}}_{[t+1, t_f]} \leq v_{\rm lat} \wedge s^{v_{\rm lat}}_{[t+1, t_f]} \geq - v_{\rm lat} ) \\
\phi_{\rm vel} =  \phi_{\rm sag} \wedge  \phi_{\rm lat}
\end{gather*}

Accordingly, the loss of the locomotion velocity specification is defined as:
\begin{equation}
    \mathcal{L}_{\phi_{\rm vel}} = \underbrace{{\rm ReLU}(-\rho((s^{v_{\rm sag}},s^{v_{\rm lat}}),\phi_{\rm vel}))}_{\textit{velocity violation}}
    \label{eq:velocity}
\end{equation}

\subsubsection{Heading change specification $\phi_{\Delta \theta}$}
Let $s^{\Delta \theta}_{[t+1]}$ be a signal equal to the heading change between $\hat{\mathcal{T}}^{\rm ego}_{t}$ and $\hat{\mathcal{T}}^{\rm ego}_{t+1}$, heading change specification is:
\begin{equation*}
\phi_{\Delta \theta} = \square_{[t+1, t_f]}(s^{\Delta \theta}_{[t+1, t_f]} < \Delta \theta_{max} \wedge s^{\Delta \theta}_{[t+1, t_f]} > -\Delta \theta_{max}) 
\end{equation*}

Therefore, the loss of the heading change specification is:
\begin{equation}
    \mathcal{L}_{\phi_{\Delta \theta}} = \underbrace{{\rm ReLU}(-\rho(s^{\Delta \theta},\phi_{\Delta \theta}))}_{\textit{heading change violation}}
    \label{eq:heading_change}
\end{equation}

\subsection{Training}

The network is trained end to end using the following loss function: $\mathcal{L} = \mathcal{L}_0 +  \mathcal{L}_{\rm STL}$, where $\mathcal{L}_0$ includes a KL divergence term, endpoint loss, and average trajectory loss, akin to the loss function utilized in~\cite{mangalam2020not}. $\mathcal{L}_{\rm STL}$ is the sum of STL specification losses $\mathcal{L}_{\rm STL} = \alpha_1 \mathcal{L}_{\phi_{\Delta \theta}} + \alpha_2 \mathcal{L}_{\phi_{\rm vel}}$, where $\alpha_1$ and $\alpha_2$ are constant weights.










\section{Bipedal Motion Planning}
\subsection{Reduced-order MPC}
At the middle level of our navigation framework in Fig.~\ref{fig:high-level}, we employ a ROM-based MPC as a step planner for the bipedal system~\cite{narkhede2022sequential}. The objective of this layer is to design a ROM motion plan that tracks the desired trajectory produced by the social path planner. 
The ROM utilized to design the motion of Digit is the Linear Inverted Pendulum (LIP) model.
\subsubsection{Model Dynamics}
For the LIP model we assume that each step has a fixed duration $T$\footnote{set to be equal to the timestep between frames in the dataset ($0.4$ s)}, the discrete sagittal dynamics\footnote{the lateral dynamics are only considered at the low level since it is periodic with a constant desired foot placement.} of the $q^{\rm th}$ walking step of the LIP model are:
\begin{equation}
   \boldsymbol{x}_{q+1} = \begin{bmatrix}
x_{q+1} \\
\dot{x}_{q+1} 
\end{bmatrix}	= \begin{bmatrix}
 \sinh(\omega T)/\omega \\
\cosh(\omega T)  
\end{bmatrix}  \dot{x}_{q}
+ \begin{bmatrix}
1 -\cosh(\omega T) \\
-\omega \sinh(\omega T)
\end{bmatrix}  u^{f}_{q} 
\end{equation}
where $u^{f}_{q}$ is the sagittal foot position relative to the CoM, $x_{q+1}$ is incremental change in CoM between walking steps (see the ALIP model in Figure.~\ref{fig:high-level}), and $\omega = \sqrt{g/H}$, where $g$ is the gravitational constant and $H$ is the CoM height. To plan motions for the LIP model in a global coordinate, we set the heading change to be $\theta_{q+1} = \theta_{q} + u^{\Delta \theta}_q$ and the global CoM positions to be:
\begin{equation}
   \boldsymbol{x}^{g}_{q+1} = \begin{bmatrix}
        x^{g}_{q+1} \\
        y^{g}_{q+1}
    \end{bmatrix} = 
    \begin{bmatrix}
        x^{g}_{q} \\
        y^{g}_{q}
    \end{bmatrix} + \begin{bmatrix}
         \cos{\theta_{q}} \\
        \sin{\theta_{q}}
    \end{bmatrix}  x_{q+1}
\end{equation}

\subsubsection{Constraints}

To prevent the LIP dynamics from taking a step that is infeasible by the Digit robot the following constraint is implemented
\begin{equation}
    \mathcal{XU}_q = \{(\boldsymbol{x}_q, \boldsymbol{u}_q) \; | \; \boldsymbol{x}^{\rm lb} \leq \boldsymbol{x}_q \leq \boldsymbol{x}^{\rm ub} \; \text{and} \; \boldsymbol{u}^{\rm lb} \leq  \boldsymbol{u}_q \leq \boldsymbol{u}^{\rm ub} \}
    \label{eq:const}
\end{equation}
where $ \boldsymbol{u}_q = [u^{f}_{q} \quad u^{\Delta \theta}_q]$, $\boldsymbol{x}^{\rm lb}$ and $\boldsymbol{x}^{\rm ub}$ are the lower and upper bound of $\boldsymbol{x}_q$ respectively, and $\boldsymbol{u}^{\rm lb}$ and $\boldsymbol{u}^{\rm ub}$ are the bounds for $\boldsymbol{u}_q$.

To enforce navigational safety at the middle level, a control barrier function constraint \cite{teng2021toward,narkhede2022sequential} is implemented to avoid the collision with the closest pedestrian $(x^{p_c},y^{p_c})$ as such:
\begin{equation*}
    h(\boldsymbol{x}^g) = \left(\left(\frac{x^g - x^{p_c}}{0.2}\right)^2+\left(\frac{y_g - y^{p_c}}{0.2}\right)^2 \right)^\frac{1}{2} -1
\end{equation*}
where $h(\boldsymbol{x}^g_0) \geq 0$.

\begin{figure*}[h]
\centerline{\includegraphics[width=.95\textwidth]{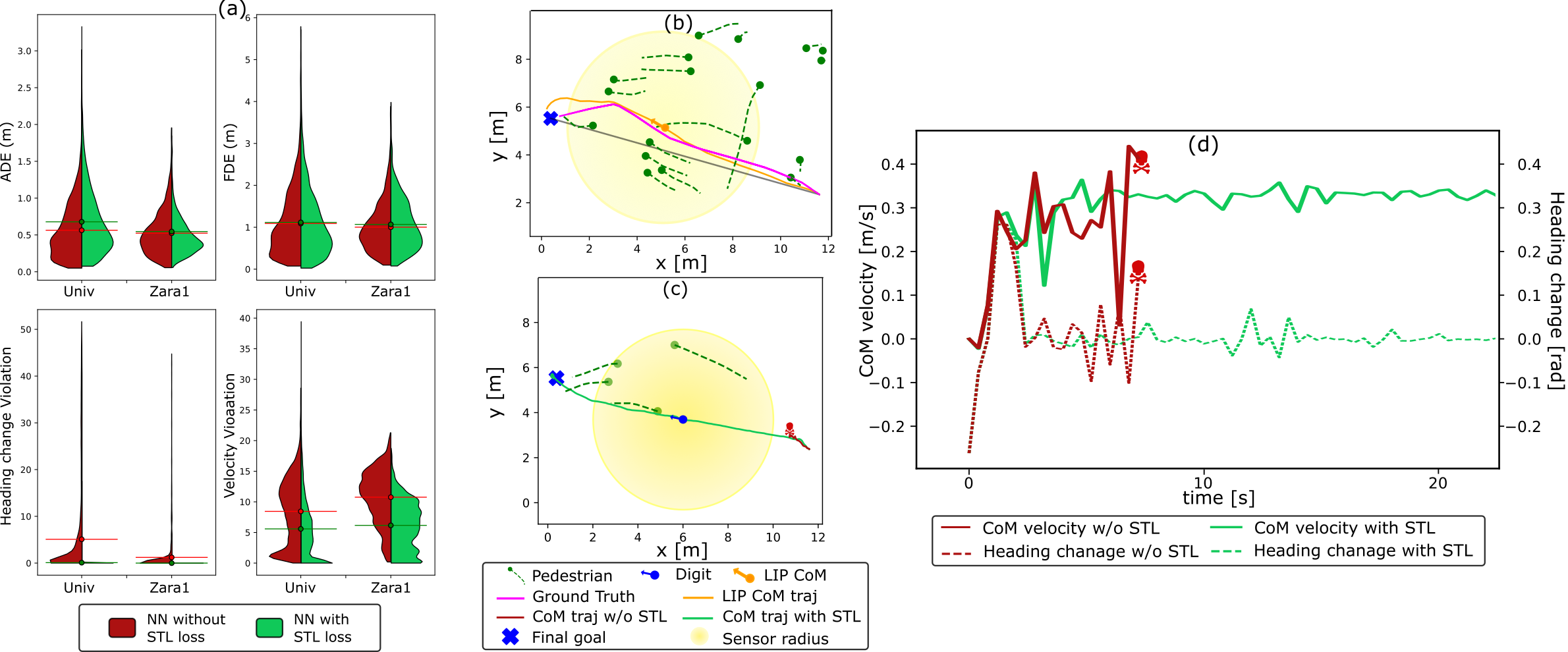}}
\caption{Results of testing two neural networks with and without STL locomotion losses are shown in green and red, respectively. (a) shows violin plots of ADE, FDE, heading change violation, and velocity violation for two different sets of data UNIV and ZARA1. (b) shows the LIP CoM trajectory along with the ground truth trajectory from the crowd dataset. (c) Shows Digit's CoM trajectory, and (d) shows commanded CoM velocity and heading change. }
\label{fig:training_results}
\end{figure*}
\subsubsection{Cost}
Let $S\mathrm{x}_{q} = [x^g_{q} \quad y^g_{q} \quad \theta_{q} \quad \dot{x}_{q}]$, the running cost is defined as:
\begin{equation}
    J_q(S\mathrm{x}_{q}) = \| S\mathrm{x}_{q} - \boldsymbol{x}_{\hat{\mathcal{T}}^{\rm ego}_{q}} \|^2_{W_1}
\end{equation}
where $\boldsymbol{x}_{\hat{\mathcal{T}}^{\rm ego}_{q}} = [\hat{\mathcal{T}}^{\rm ego}_{q} \quad s^{\Delta \theta}_{q} \quad  s^{v_{\rm sag}}_{q}] $ and $W_1$ is diagonal weighting matrix. The terminal cost of the MPC is defined as:
\begin{equation}
    J_N(S\mathrm{x}_{N}) = \| S\mathrm{x}_{N} - \boldsymbol{x}_{\mathcal{G}} \|^2_{W_2}
\end{equation}
where $N$ is the number of walking steps within the MPC horizon, $\boldsymbol{x}_{\mathcal{G}} = [\mathcal{G} \quad \theta_{\rm terminal} \quad \dot{x}_{\rm terminal}]$, and $W_2$ is diagonal weighting matrix.

Let $S\mathrm{x}_{q+1} = \Phi(S\mathrm{x}_q, \boldsymbol{u}_q)$ be the dynamics describing the evolution of $S\mathrm{x}$. Similar to \cite{narkhede2022sequential}, the step planner can be solved by:
\begin{equation}
\begin{aligned}
\min_{X} \quad \sum_{q=0}^{N-1} & J_q(S\mathrm{x}_q) +  J_N(S\mathrm{x}_{N}) \\
\textrm{s.t.} \quad & S\mathrm{x}_{q+1} = \Phi(S\mathrm{x}_q, \boldsymbol{u}_q)\\
  &   (\boldsymbol{x}_q, \boldsymbol{u}_q) \in \mathcal{XU}_q   \\
  &  h(S\mathrm{x}_{q+1}) \geq (1-\gamma)  h(S\mathrm{x}_{q}) \\ 
\end{aligned}
\end{equation}

\subsection{Low-level Full-Body Control}
At the low level we utilize the Angular momentum LIP planner introduced in~\cite{Gong2022AngularMomentum}, and Digit's passivity controller with ankle actuation introduced in our previous work~\cite{shamsah2023integrated}. Here we set the walking step time and lateral step width to be fixed at $0.4$ s and $0.4$ m, respectively.


\section{results and implementation}

\subsection{Implementation}

The social path planner module introduced in Sec.~\ref{sec:social_path_planner} was trained on the UCY~\cite{lerner2007crowds} and ETH~\cite{pellegrini2009walk} crowd datasets with the common leave-one-out approach, reminiscent of prior studies~\cite{salzmann2020trajectron++,mangalam2020not, gupta2018social, li2019conditional}. To evaluate the performance of adding robot locomotion specifications into the training, we trained two neural network models with and without the added robot-specific losses introduced in Sec.~\ref{subsec:stl_loss}. These models were trained on two distinct sets of datasets, with one set excluding UNIV from the training examples, and the other set excluding ZARA1.
We employ a historical trajectory observation $\mathcal{T}^{p_k}_{[t_p, t]}$ and a prediction horizon $\hat{\mathcal{T}}^{\rm ego}_{[t, t_f]}$, each spanning a duration of $8$ timesteps ($3.2$ s) and only consider neighboring pedestrians that are within a radius of $4$ m.

\subsection{Evaluating Social Prediction Accuracy with Bipedal Locomotion Safety Specifications}


In Figure.~\ref{fig:training_results}, We show that the standard prediction accuracy metrics (ADE and FDE) are minimally affected by the addition of the locomotion-specific losses\footnote{Anticipated accuracy deviations may arise from variances in human locomotion, including abrupt halts, turns, and differing walking speeds that may not align with Digit's safe operational parameters. } as seen in Figure.~\ref{fig:training_results}(a), while the heading change violation and velocity violation are significantly reduced in the social planner model that includes the locomotion safety specification losses. 

As our objective is to employ the prediction module as a robot path planner, the focus remains on consistently identifying viable walking paths. Our intention does not encompass predicting the ego-agent's trajectory during stationary periods or when the ground truth trajectory deviates towards an interim route apart from the ultimate destination $\mathcal{G}$. In such scenarios, elevated ADE and FDE values emerge.


\subsection{Digit Social Navigation Task}
First, we evaluate our framework by substituting one pedestrian with the LIP model and generating motion plans at the middle level. Figure.~\ref{fig:training_results}(b) shows the LIP CoM trajectory (orange) compared to the substituted pedestrian ground truth trajectory (magenta).

Second, we evaluate the locomotion safety by employing identical parameters and initial conditions for both the MPC and the low-level controller, we empirically demonstrate that Digit consistently falls when using the social path planner without the STL-based locomotion safety losses as seen in Figure.~\ref{fig:training_results}(c). Figure.~\ref{fig:training_results}(d) shows the CoM velocity and the heading change produced by the social path planner corresponding to the trajectories illustrated in Figure.~\ref{fig:training_results}(c). The CoM velocity and the heightened rate of heading change produced by the social path planner without the STL-based locomotion safety losses result in failure.
\begin{rem}
    Certainly, while using overly conservative constraints in MPC (Eq.(\ref{eq:const})) can ensure safe locomotion, employing the social path planner with STL-based safety permits less conservative constraints without compromising safety. 
\end{rem}


\section{Conclusion}
We presented a hierarchical planning framework designed for socially acceptable navigation of bipedal robots. Our framework generates a socially acceptable path while maintaining the locomotion safety of the bipedal system. Future work will include; (i) enhancing STL specification design to encompass a wide range of safety parameters in robot navigation; 
(ii) a middle-level locomotion planner that factors in the anticipated trajectories of surrounding pedestrians; (iii) in-depth comparisons with other human trajectory prediction modules; and (iv) comparing the success rates of the proposed framework with and without the incorporated locomotion safety losses.






\bibliographystyle{IEEEtran}
\bibliography{lidar.bib}

\end{document}